\documentclass{article}

\usepackage{microtype}
\usepackage{graphicx}
\usepackage{subfigure}
\usepackage{booktabs} 

\usepackage{hyperref}

\usepackage[accepted]{icml2024}

\usepackage{amsmath}
\usepackage{amssymb}
\usepackage{mathtools}
\usepackage{amsthm}

\usepackage[capitalize,noabbrev]{cleveref}

\theoremstyle{plain}

\theoremstyle{definition}

\theoremstyle{remark}

\usepackage{outlines}
\usepackage{makecell}
\usepackage{caption}
\usepackage{subcaption}

\newcommand{\methodshort}{SSO}
\newcommand{\methodlong}{Skill Set Optimization}
\newcommand{\algcomment}[1]{\textit{\textcolor{gray}{// #1}}}
\DeclareMathOperator{\E}{\mathbb{E}}

\renewcommand{\algorithmiccomment}[1]{\hfill$\triangleright$\textit{#1}}

\icmltitlerunning{\methodlong{}}

\begin{document}

\twocolumn[
\icmltitle{\methodlong{}:\\Reinforcing Language Model Behavior via Transferable Skills}

\icmlsetsymbol{equal}{*}

\begin{icmlauthorlist}
\icmlauthor{Kolby Nottingham}{uci}
\icmlauthor{Bodhisattwa Prasad Majumder}{equal,ai2}
\icmlauthor{Bhavana Dalvi Mishra}{equal,ai2}\\
\icmlauthor{Sameer Singh}{uci}
\icmlauthor{Peter Clark}{ai2}
\icmlauthor{Roy Fox}{uci}
\end{icmlauthorlist}

\icmlaffiliation{uci}{Department of Computer Science, University of California Irvine, Irvine CA, United States}
\icmlaffiliation{ai2}{Allen Institute for AI, Seattle Washington, United States. * Equal contribution}

\icmlcorrespondingauthor{Kolby Nottingham}{knotting@uci.edu}

\icmlkeywords{language models, natural language processing, decision making, reinforcement learning}

\vskip 0.3in
]

\printAffiliationsAndNotice{}

\begin{abstract}
    Large language models (LLMs) have recently been used for sequential decision making in interactive environments.
    However, leveraging environment reward signals for continual LLM actor improvement is not straightforward.
    We propose \methodlong{} (\methodshort{}) for improving LLM actor performance through constructing and refining sets of transferable skills.
    \methodshort{} constructs skills by extracting common subtrajectories with high rewards and generating subgoals and instructions to represent each skill.
    These skills are provided to the LLM actor in-context to reinforce behaviors with high rewards.
    Then, \methodshort{} further refines the skill set by pruning skills that do not continue to result in high rewards.
    We evaluate our method in the classic videogame NetHack and the text environment ScienceWorld to demonstrate \methodshort's ability to optimize a set of skills and perform in-context policy improvement.
    \methodshort{} outperforms baselines by 40\% in our custom NetHack task and outperforms the previous state-of-the-art in ScienceWorld by 35\%.
\end{abstract}

\begin{figure}
    \centering
    \includegraphics[width=.85\linewidth]{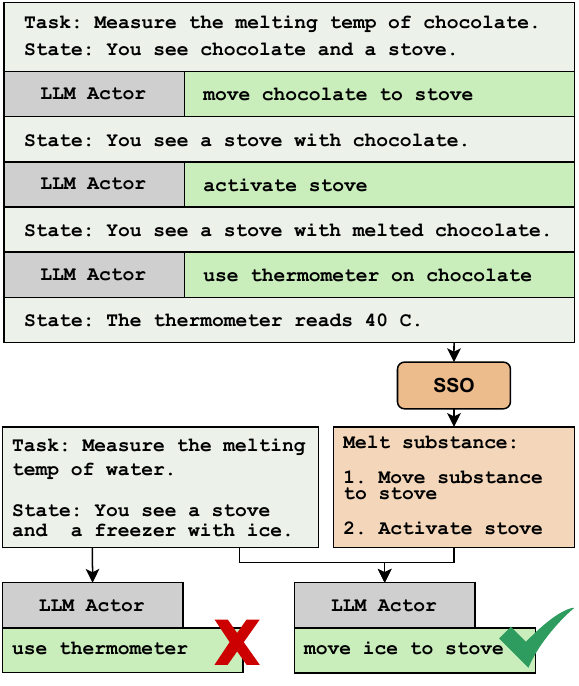}
    \caption{
        Example of a interactive text task and skill.
    }
    \label{fig:example}
    \vspace{-.5cm}
\end{figure}

\begin{figure*}
    \centering
    \includegraphics[width=\linewidth]{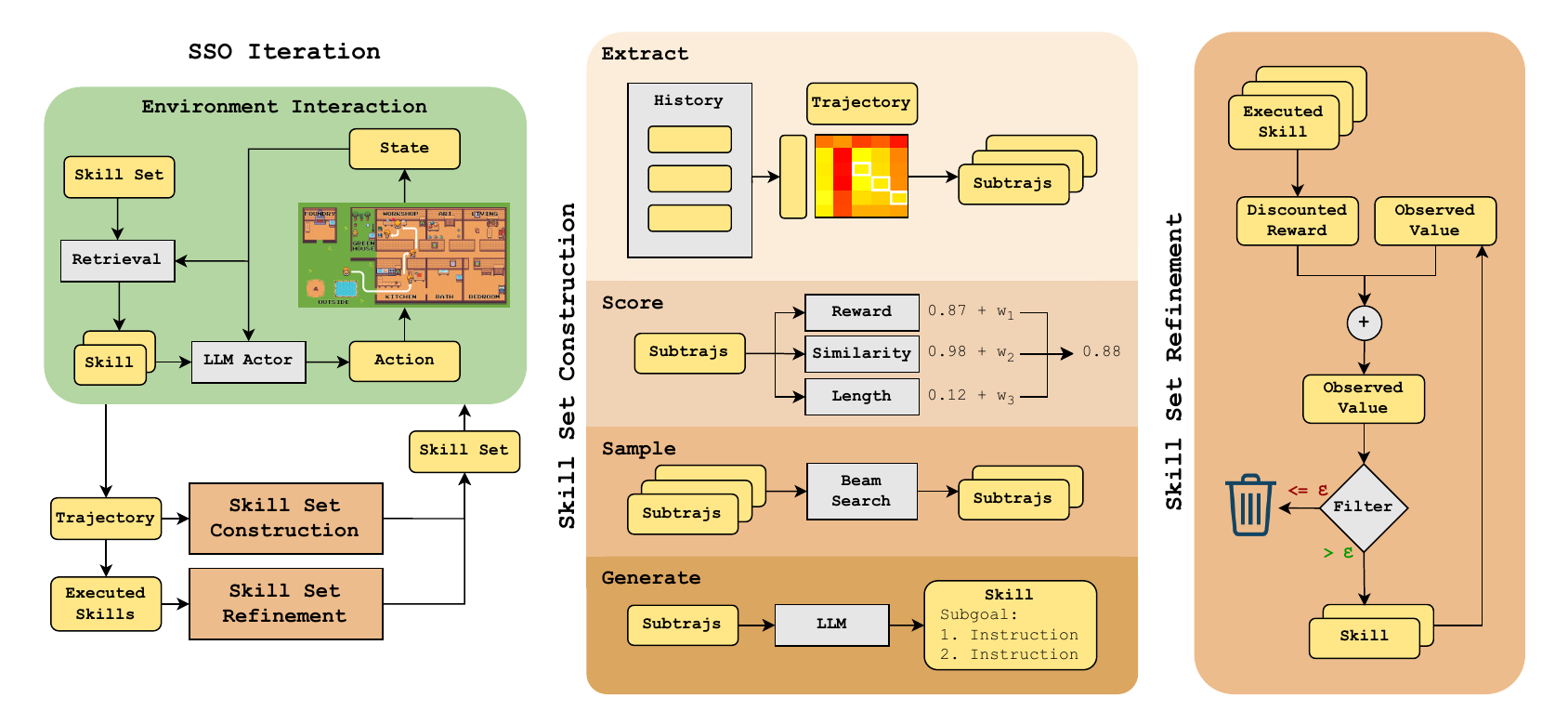}
    \caption{
        Each iteration of \methodshort{} collects a trajectory of interactions with the current LLM actor, uses this trajectory to construct new skills and filter poorly performing skills, and updates the skill set for use in the next iteration.
        New skills are constructed by extracting, scoring, and sampling sets of similar subtrajectories that are then used to generate subgoals and instructions for skills.
        Skills are filtered based on the discounted future rewards observed after executing a skill.
    }
    \label{fig:sso}
\end{figure*}

\section{Introduction}

Large Language Model (LLM) actors have been deployed in interactive domains such as robotics \cite{ichter2022do,huang2022language,huanginner}, games \cite{nottingham2023embodied,nottingham2023selective}, and programming \cite{chen2022codet}.
Similar to the reinforcement learning (RL) setting, these domains often provide a reinforcement signal in the form of reward, task success, or user feedback.
For example, the task in Figure \ref{fig:example} may provide a success signal upon measuring the substance's temperature and intermediate rewards for completing subgoals such as activating the stove.
However, finetuning an LLM actor directly using a traditional RL policy gradient is often impractical with contemporary LLMs and impossible with black-box closed-source LLMs.
Instead, we explore a new paradigm of in-context policy improvement.

In natural language processing (NLP) tasks, in-context learning improves task performance by editing LLM inputs with instructions \cite{brown2020langauge}, task examples \cite{wei2021finetuned}, or auxiliary tasks \cite{wei2022chain}.
However, interactive domains require sequential decision making with long trajectories and complex credit assignment (see Section \ref{sec:incontext}), so naively applying in-context learning techniques generalizes poorly and does not scale well.
Instead, recent approaches for improving LLM actors construct a ``memory'' of knowledge about the world \cite{majumder2023clin}, skills \cite{wang2023voyager}, or task insights \cite{shinn2023reflexion} to use in-context for policy improvement.

Current approaches to constructing ``memory'' for LLM actors have shortcomings such as a lack of continual memory evaluation, intermediate subgoals, and memory retrieval.
To address these shortcomings, we propose \textbf{\methodlong{}} (\textbf{\methodshort{}})\footnote{\href{https://allenai.github.io/sso/}{https://allenai.github.io/sso/}}, a method for automatically constructing skills for in-context policy improvement, where a skill is composed of an initial state, a list of language instructions, and a language subgoal.
\methodshort{} takes inspiration from both in-context learning and policy optimization to optimize a set of skills for in-context policy improvement.
Figure \ref{fig:sso} shows how skills are constructed from past trajectories by extracting similar subtrajectories, scoring and sampling sets of subtrajectories that do not overlap, and finally generating an abstract subgoal and instructions for each set of subtrajectories.
The constructed skill set is later refined by filtering skills that do not lead to high future rewards when executed.
By prioritizing skills according to environment reward at the construction and refinement steps, we identify subgoals that have high impact on policy improvement.
Additionally, the subtrajectories' initial states provide a reference for skill retrieval allowing \methodshort{} to retrieve skills that are immediately relevant.
Finally, because each skill subgoal and instructions are generated from multiple subtrajectories, the resulting subgoals and instructions are often abstract (as in Figure \ref{fig:example}) and further facilitate task transfer.

We empirically demonstrate the advantage of using skills for in-context policy improvement and evaluate \methodshort's ability to rapidly adapt and transfer skill sets in two sequential decision making domains.
First, we evaluate \methodshort{} on the text-based environment ScienceWorld \cite{wang2022scienceworld} and on the grid-based game NetHack \cite{kuttler2020nle}.
\methodshort{} achieves state-of-the-art performance on ScienceWorld, outperforming the previously top performing CLIN agent \cite{majumder2023clin} by an average of 35\%.
\methodshort{} also outperforms baselines on our custom NetHack task by 40\% despite its low-level action space.
Our analysis and ablations show that \methodshort{} continually optimizes an LLM actor's policy by extracting increasingly helpful skills that maximize the task's reinforcement signal.

\section{Related Work}

\subsection{In-Context Learning}

Finetuning state-of-the-art LLMs can be prohibitively expensive.
In-context learning utilizes an LLM's ability to recognize patterns by augmenting prompts to modify LLM behavior \cite{brown2020langauge,wei2021finetuned,wei2022chain}.
Instruction tuning facilitates this by finetuning LLMs to follow custom instructions \cite{brown2020langauge}.
Previous work has attempted to learn prompts that maximize performance on a dataset \cite{deng-etal-2022-rlprompt,zhang2023tempera,fernando2023promptbreeder}.
However, to maintain interpretability, these methods typically merely edit existing prompts. 
Also, they require constant evaluation on a downstream task which is often impractical for sequential tasks with delayed rewards.

\subsection{LLMs for Sequential Decision Making}

LLMs are popular tools for planning and high-level decision making in robotic applications \citep{ichter2022do,huang2022language,huanginner} games \citep{nottingham2023embodied,nottingham2023selective}, programming \cite{chen2022codet}, and computer tasks \cite{kim2023language,liang2023taskmatrix}.
Previous work focuses on improved prompting and evaluation methods for LLM actors without considering approaches for continual learning.
This is partly because continual learning for LLMs via finetuning is very expensive and often impractical, especially in sequential decision making domains with long contexts and noisy credit assignment.
However, in this paper, we leverage in-context learning for sequential decision making to develop a powerful method for quickly adapting to and generalizing between tasks.

\subsection{LLMs with Environment Feedback}

Some recent work has leveraged task success signals from an environment for in-context policy improvement.
The Reflexion \cite{shinn2023reflexion}, Voyager \cite{wang2023voyager}, DEPS \cite{wang2023deps}, and ADAPT \cite{prasad2023adapt} agents attempt to retry tasks or subgoals after a failure.
These methods work by prompting an LLM to reflect on the failed attempt and suggest improvements to make on the next attempt.
However, none of these methods, with the exception of Voyager (see below), leverage successful task completions for learning and are not compatible with task transfer or generalization.

This paper focuses on the problem of building a memory-like collection of helpful information for adaptation to new tasks and transfer between tasks.
Previous work that has pursued this research direction include the Voyager \cite{wang2023voyager}, ExpeL \cite{zhao2023expel}, and CLIN \cite{majumder2023clin} agents.
Voyager generates javascript skills, self-corrects mistakes with a reflexion-like process, and then stores all successful skills in-context.
Similar to \methodshort{}, Voyager's skills store instructions for reaching a subgoal.
However, Voyager's skills are executed as code and explore future subgoals that are yet to be reached rather than extracting subgoals from past experience to maximize reward.
ExpeL prompts an LLM to generate or edit free-form insights based on successful trajectories.
Unlike \methodshort{}, ExpeL does not leverage partially successful trajectories when generating insights.
CLIN generates a memory of insights in the form ``action X [may/should] be necessary to do action Y'' by prompting an LLM with past trajectories.
Like \methodshort{}, CLIN does not require a past trajectory to be successful in learning useful information, but CLIN does not take task performance or policy improvement into account when constructing its memory.
Additionally, no previous method continually evaluates memories based on environment feedback like in \methodshort's skill refinement step.

Most of the methods for in-context policy improvement in this section are complementary.
For example, the ExpeL agent utilizes reflexion and fewshot trajectories in addition to its memory of insights. 
However, for the sake of making a direct comparison, our agent only uses \methodshort{} for policy improvement, and we choose to compare to one self-correcting method (Reflexion), one insight method (CLIN), and fewshot example trajectories.

\section{In-Context Policy Improvement}
\label{sec:incontext}

With the increased dominance of LLMs in NLP, in-context learning has become an essential tool for improving performance on NLP tasks.
Rather than doing expensive finetuning on downstream tasks, in-context learning enables significant increases in task performance just by adding supplementary inputs to the LLM.
In-context changes to an LLM actor may be an efficient way to make continual changes to the actor's policy.
However, previous work has done little to explore effective methods of leveraging environment feedback for LLM actor in-context learning.

Unlike other NLP tasks, sequential decision making in an interactive environment requires multiple outputs from an LLM actor in the correct sequence to generate a successful trajectory.
Sequential decision making is often represented as a Markov Decision Process (MDP) with states $s\in\mathcal{S}$, actions $a\in\mathcal{A}$, a transition propability function $T:\mathcal{S}\times\mathcal{A}\times\mathcal{S}'\to[0,1]$, and a reward function $R:\mathcal{S}\times\mathcal{A}\to\mathbb{R}$.
In the case of an LLM actor, environment states are the text inputs to the LLM, and the output of the LLM is executed as an action in the environment.
The objective of an LLM actor is to model an optimal policy $\pi(a|s)$ that maximizes rewards in a trajectory $\tau={s_0,a_0,...,s_T,a_T}$:

$$J(\pi_\theta)=\underset{\tau \sim \pi_\theta,T}{\E} \bigg[ \sum_{t=0}^{|\tau|} R(s_t, a_t) \bigg]$$

To avoid traditional gradient-based policy optimization techniques, which can be computationally prohibitive with LLMs or impossible with black-box LLMs, we assume that the policy $\pi$ is parameterized by text inputs $\theta$ that are provided to the agent alongside the state $s$.
A successful in-context policy improvement method will identify a $\theta$ that increases the objective $J(\pi_\theta)$.

A straightforward adaptation of in-context learning for in-context policy improvement would be to provide an LLM actor with examples of successful trajectories, thus providing information to the LLM actor about the optimal policy.
However, this approach does not scale well as it results in long context lengths and may include redundant information or information that does not transfer well between tasks. 
Instead, we develop a scalable and transferable memory structure for in-context policy improvement that we call skills.
Skills leverage the sequential information from trajectory examples while keeping information brief and modular.
A skill is composed of an initial state for used for retrieval, a final state used as a natural language subgoal, and a sequence of actions used as natural language instructions for reaching that subgoal.

\begin{figure}
    \centering
    \includegraphics[width=\linewidth]{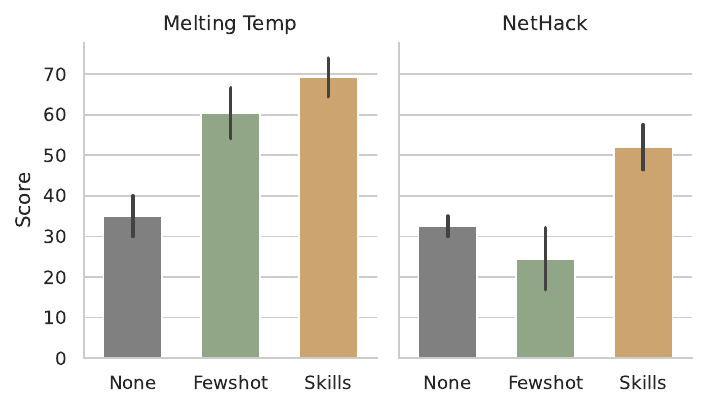}
    \caption{
    Comparison between in-context skills, fewshot trajectory examples, and no in-context information on the Melting Temperature ScienceWorld and NetHack tasks.
    }
    \label{fig:fewshot}
\end{figure}

To demonstrate the difficultly of using fewshot trajectory examples for in-context policy improvement, we compare the use of fewshot trajectories vs. skills on ScienceWorld \cite{wang2022scienceworld} and NetHack \cite{kuttler2020nle} domains (see Section \ref{sec:setup}).
For the fewshot LLM actor, we gather 30 trajectories using an LLM actor and select the three best trajectories to provide in-context at each step.
For the skill-based LLM actor, we use the same 30 trajectories to create skills using \methodshort{}, and we retrieve three skills to provide in-context at each step.
Figure \ref{fig:fewshot} shows how well skills perform vs. fewshot examples for in-context policy improvement.
This is especially the case for the NetHack domain where trajectories are longer and actions are expressed characters instead of language actions.
Also, the context for the fewshot actor was on average 5x longer than the context for the skill-based actor.
The following section ellaborates on how we use \methodshort{} to learn and optimize a set of skills for continual in-context policy improvement.

\section{\methodlong{}}
\label{sec:sso}

We propose optimizing LLM actors by providing transferable skills in-context.
First, we construct new skills for our skill set from subtrajectories with high rewards to reinforce successful behaviors.
Second, we further refine the constructed skill set by filtering skills that do not result in high rewards when used in subsequent trajectories.
When using skills in-context, we retrieve relevant skills to include in context based on cosine similarity of skill initial states and the current environment state.
Each iteration of \methodshort{} includes rolling out a single trajectory with the current LLM actor and skills, constructing new skills, and filtering out skills that did not result in positive rewards in the last trajectory.
The updated skill set is then used by the LLM actor in the following iteration.
This process is illustrated in Figure \ref{fig:sso} and described in the following sections.

\subsection{Skill Set Construction}

A skill is expressed to the actor as an abstract LLM-generated subgoal and list of instructions for reaching that subgoal.
We define a unique skill using one or more subtrajectories where the final states in each of the subtrajectories are used to generate the subgoal and the actions in each subtrajectory are used to generate the instructions for reaching that subgoal.
Given the LLM actor's previous trajectories, we \textbf{extract} potential subtrajectories, \textbf{score} them using several heuristics, \textbf{sample} a skill set using beam search, and \textbf{generate} subgoals and instructions for each skill.

\subsubsection{Extract}

To identify transferable skills, we extract multiple similar subtrajectories to be used to generate each skill.
The subtrajectories of a skill must be similar enough that a common abstract subgoal and instructions can be generated.
We estimate the similarity of two subtrajectories using the average cosine similarity of each of their state and action embeddings.
Using multiple subtrajectories for each skill has two important benefits:
(1)~the resulting skills are abstract and more transferable, and
(2)~repeated similar subtrajectories is a strong signal for identifying useful subgoals.

Enumerating every possible set of subtrajectories from an LLM actor's experience would be infeasible, so we only consider pairs of subtrajectories of certain lengths ([2,5] in our experiments) that come from different trajectories.
After each completed trajectory, we enumerate its subtrajectories and extract the most similar subtrajectory from each of the $N$ previous completed trajectories, as shown in Algorithm \ref{alg:extract}. Each of these subtrajectory pairs are considered for potential skills, although most would not result in a useful skill.

\begin{algorithm}[h]
\caption{Extract}\label{alg:extract}
\begin{algorithmic}
    \REQUIRE $\mathcal{D}$ \hfill\algcomment{$N$ past trajectories}
    \REQUIRE $\tau$ \hfill\algcomment{latest trajectory}
    \REQUIRE $min, max$ \hfill\algcomment{min and max skill length}
    \STATE $SS \gets \emptyset$ \hfill\algcomment{skill set}
    \STATE\algcomment{iterate subtrajectories of length [$min$,$max$]}
    \FOR{$\tau_{sub} \in GetAllSubtraj(\tau, min, max)$}
        \FOR{$\hat{\tau} \in \mathcal{D}$}
            \STATE\algcomment{get the most similar subtrajectory of same length}
            \STATE $\hat{\tau}_{sub} \gets SimilarSubtraj(\tau_{sub}, \hat{\tau})$
            \STATE $SS \gets SS \cup \{(\tau_{sub}, \hat{\tau}_{sub})\}$
        \ENDFOR
    \ENDFOR
\end{algorithmic}
\end{algorithm}

\subsubsection{Score}

The majority of extracted subtrajectory pairs must be removed from canidacy for the skill set.
Pairs should be similar enough to have common subgoal and instructions, lead to high rewards, and have high coverage of past experience.
To accomplish this, we calculate the average state and action similarity, discounted future reward, and length of each pair.
We compute a score using a weighted sum of these values to identify which subtrajectory pairs would make useful skills.
In our experiments, we set score weights to $w_1=1$, $w_2=0.1$, and $w_3=0.01$ to prioritize first similarity, then reward, and finally length.

\subsubsection{Sample}

\begin{figure*}[t!]
  \begin{minipage}[b]{0.65\linewidth}
    \centering
    \small
\begin{tabular}{l|c|ccc|cc}
\toprule
ScienceWorld        &      & \multicolumn{3}{c|}{Adaptation} & \multicolumn{2}{c}{Transfer}        \\
Task                &ReAct &Reflexion   &CLIN         & SSO          & CLIN         & SSO          \\
\midrule
Temperature         & 7.2  & 5.9        & 14.3        & \textbf{100} & 15.7         & \textbf{71.6}\\
Melting Temp        & 6.1  & 28.6       & 51.8        & \textbf{97.3}& 49.7         & \textbf{69.2}\\
Find Plant          & 26.7 & 64.9       &\textbf{100} & \textbf{100} & 59.2         & \textbf{100} \\
Find Living         & 53.3 & 16.4       &\textbf{100} & 96.7         & \textbf{100} & 90           \\
Chemistry           & 51   & 70.4       & 44.4        & \textbf{82.6}& 42.2         & \textbf{48}  \\
Color Mixing        & 58.9 & 70.7       & 56.7        & \textbf{81.1}& \textbf{85.6}& 71.1         \\
Lifespan, Longest   & 61   &\textbf{100}&\textbf{100} & \textbf{100} & 65           & \textbf{90}  \\
Lifespan, Shortest  & 67.5 & 84.4       & 90          & \textbf{100} & 75           & \textbf{80}  \\
Life Stages, Plant  & 8    & \textbf{8} & \textbf{8}  & 6.2          & \textbf{32}  & 3.4          \\
Life Stages, Animal & 27.7 & 2.6        & 81          & \textbf{100} & 42.8         & \textbf{77}  \\
Boil                & 3.5  & 4.2        & 15.2        & \textbf{81.7}& 4.4          & \textbf{48.7}\\
Freeze              & 7.8  & 7.8        & 10          & \textbf{74.3}& 8.9          & \textbf{38.9}\\
Grow Plant          & 9.1  & 7.3        & 11          & \textbf{86.6}& 10.9         & \textbf{61.2}\\
Grow Fruit          & 18.6 & 13         & 71.6        & \textbf{78}  & \textbf{70.8}& 28.3         \\
Gravity             & 40.5 & 50.6       &\textbf{100} & \textbf{100} & 70           & \textbf{74}  \\
Friction            & 44   &\textbf{100}& 72.5        & 94           & \textbf{70}  &  67.5        \\
Genetics, Known     & 25.7 & 50.9       &\textbf{100} & 78.5         & \textbf{84.5}&  42.5        \\
Genetics, Unknown   & 16.8 & 23.7       &\textbf{92.6}& 48.7         & \textbf{61.4}&  20.3        \\
\midrule
Average             & 29.6 & 39.4       & 62.2        & \textbf{83.7}& 52.7         & \textbf{60.1}\\
\bottomrule
\end{tabular}
  \end{minipage}
  \hfill
  \begin{minipage}[b]{0.37\linewidth}
      \begin{minipage}[b]{\linewidth}
        \includegraphics[trim={-1.5cm -.5cm 0 -.5cm},width=.83\linewidth]{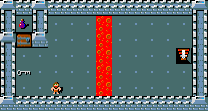}
      \end{minipage}
      \begin{minipage}[b]{\linewidth}
        \includegraphics[trim={0 6cm 0 0},width=.85\linewidth]{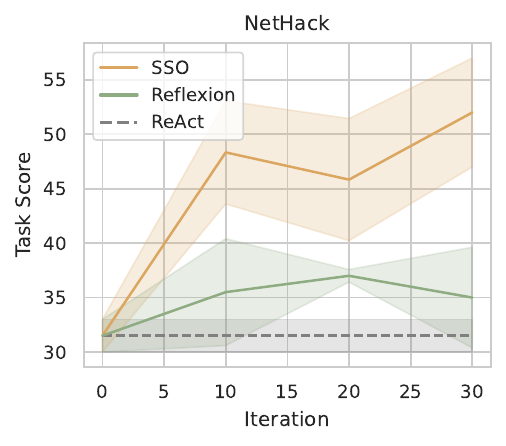}
      \end{minipage}
  \end{minipage}
  \caption{
    We compare \methodshort{} with ReAct and Reflexion baselines in ScienceWorld and NetHack domains.
    We also compare with the previous state-of-the-art for ScienceWorld, CLIN.
    In ScienceWorld we evaluate adaptation---attempting a single variant five times---and transfer---learning on 10 train variants for 30 iterations before testing on the heldout test variants.
    In NetHack we test task adaptation across 30 iterations.
  }
  \label{fig:res}
\end{figure*}

We utilize subtrajectory pair scores to conduct a beamsearch over pairs to maximize the sum of scores.
Many of the extracted potential skills have significant overlap with each other, so we impose a constraint on the beamsearch that there can be no overlap of skill subtrajectories in a skill set.
We also include all subtrajectory pairs in the search that were sampled during previous iterations of \methodshort{}.
The final sampled skill set includes unique skills that prioritize similar, high reward, multi-step subtrajectory pairs.

\subsubsection{Generate}

Finally, the subtrajectories in each pair are summarized into a single subgoal and list of instructions.
We prompt the actor LLM with the pair of subtrajectories and ask it to generate ``a numbered list of instructions for completing the skill'' and ``a single target observation that would indicate the success of the skill''.
We also ask the LLM to remove any semantically identical skills that may remain.
Note that this is the only step in \methodshort{} that utilizes a generative-LLM.
We found that depending on generative-LLMs for extraction and scoring was less reliable in addition to increasing costs.

\subsection{Skill Set Refinement}
\label{sec:refine}

The previously described skill set construction prioritizes including transferable high-reward skills the skill set.
Experimentally, including constructed skills in context improves action accuracy across each subgoal in Table \ref{tab:skill_eval}.
We further propose the following method for refining the skill set by evaluating environment rewards after executing a skill.

To evaluate skill performance, we record the discounted future return after executing a particular skill.
Since it is not straightforward to identify which skill is being executed at each step, we ask the LLM actor to self-report when it is executing a skill with the prompt ``\textit{output which of the given subgoals you are targeting next and
then output the next action to execute}''.
Table \ref{tab:skill_eval} reports correctly self-reports skill execution about 70\% of the time.
Each time the LLM actor self-reports using a skill, we compute the discounted sum of future rewards and add this to the skill's ``observed value''.
If a skill's observed value is ever below or equal to a threshold $\epsilon$ (zero in our experiments), we filter that skill out of the skill set.
This process refines our skill set to include skills that demonstrably result in high rewards.
Example code for self-refinement can be found in Appendix \ref{app:code}.

\begin{table}[h]
    \small
    \setlength{\tabcolsep}{2.75pt}
    \center
    \begin{tabular}{l|cc|c}
    \toprule
    \bf Subgoal         & \bf w/o Skill & \bf w/ Skill & \bf Self-Reporting \\
    \midrule
    Fill container  & 0.07      & 0.37     & 0.69 \\
    Heat substance  & 0.04      & 0.22     & 0.68 \\
    Mix ingredients & 0.30      & 0.36     & 0.71 \\
    \bottomrule
    \end{tabular}
    \caption{Action accuracy from example trajectories with and without learned skills in-context, and LLM actor success rate self-reporting that it is executing the provided skill.}
    \label{tab:skill_eval}
\end{table}

\section{Experimental Setup}
\label{sec:setup}
 
\subsection{Science World}
\label{sec:sienceworld}

We evaluate \methodshort's ability to quickly adapt to and transfer knowledge between tasks in the ScienceWorld domain \cite{wang2022scienceworld}.
ScienceWorld is a text-based simulator that tests common sense and reasoning.
It is organized into various tasks with train and test variants of each.
For each of the 18 task classes listed in Figure \ref{fig:res}, we test \methodshort{} and baselines on 7 to 10 test variants.
For example, the \textit{Melting Temp} task requires the LLM actor to measure the melting temperature of a substance, but what that substance is and the environment setup will be different in each variation.

We evaluate on two training modes: adaptation and transfer.
When evaluating adaptation, we allow 5 attempts on each test variant with the ability to learn between each trial.
When evaluating transfer, we train the LLM actor on 10 training variants for 30 episodes (3 trajectories each).
After training, we evaluate on the same test variants that we used to evaluate adaptation.
The environment provides intermediate rewards for completing subtasks.
A successfully completed task will have a final cumulative reward, or score, of 100.

We compare with the following GPT-4 based methods:

\begin{itemize}
    \item[] \textbf{ReAct} \cite{yao2022react} prompts an LLM actor to reason about the task before outputting an action. All other methods also incorporate this in their prompts.
    \item[] \textbf{Reflexion} \cite{shinn2023reflexion} prompts an LLM actor to reflect on failed task attempts and then retry the task.
    \item[] \textbf{CLIN} \cite{majumder2023clin} reflects on past experience to learn transition information of the form ``$\mathcal{A}$ [may/should] be necessary to $\mathcal{A}$''.
\end{itemize}

\subsection{NetHack}
\label{sec:nethack}

NetHack is a grid-based videogame that requires challenging exploration and problem solving \cite{kuttler2020nle}.
Unlike ScienceWorld, NetHack requires low-level navigation actions.
We choose to include NetHack in our evaluation because it poses a potential challenge for our method.
\methodshort{} requires aligning common sequences of states and actions when extracting skills.
However, in environments with lower-level actions such as NetHack, dissimilar action sequences can be used to achieve the same subgoal.
Additionally, NetHack uses character-based actions instead of natural language (``k'' moves the player north and ``,'' picks up an item).
Despite this, \methodshort{} is able to successfully learn helpful skills in NetHack.

We utilize the MiniHack library \cite{samvelyan1minihack} to design a custom level that tests an LLM actor's ability to explore and learn several skills to complete a task.
Figure \ref{fig:res} shows the layout of our custom task.
The LLM actor must pick up and use a key to unlock a door, pick up and use an item to begin levitating, and safely cross the lava to the goal.
The task provides intermediate rewards for achieving each subgoal.
As in ScienceWorld, task completion results in a score of 100.
Starting locations of the actor, items, and staircase are randomized in each episode so that \methodshort{} cannot learn memorized paths between subgoals.
Also, the item that is used to levitate may be a potion or a magic ring, each of which requires different actions to activate.

In our custom NetHack task, we compare \methodshort{} with ReAct and Reflexion baselines. We omit CLIN for this task because CLIN was specifically designed for ScienceWorld and reimplementing CLIN for a new domain would be non-trivial.
Similar to adaptation on ScienceWorld, we allow Reflexion and \methodshort{} to adapt to our NetHack task.
However, unlike the adaptation setup in ScienceWorld, we run these methods for 30 iterations and evaluate every 10 iterations by attempting the task 10 times with a frozen set of skills/reflections.
All LLM actors for NetHack utilize GPT-4-Turbo instead of GPT-4 to save on costs, but we found performance between the two LLMs to be similar.

\begin{figure*}[t!]
    \centering
    \includegraphics[trim={0 0 0 0},width=\linewidth]{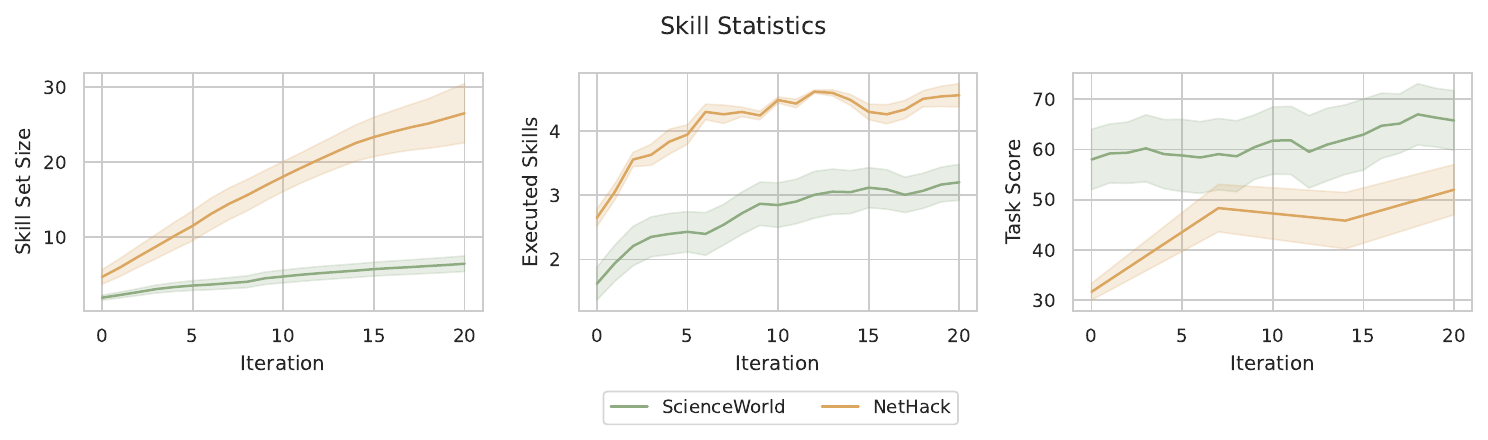}
    \caption{
    Skill Set statistics for ScienceWorld and NetHack during training.
    \emph{Skill set size} measures the number of skills created minus those that were pruned during refinement.
    \emph{Executed skills} is the average number of unique skills executed in a trajectory as reported by the LLM actor.
    Finally, the average \emph{task score} is reported throughout training.
    }
    \label{fig:stats}
\end{figure*}

\begin{table}[b!]
    \centering
    \small
    \begin{tabular}{l}
    \toprule
    \bf ScienceWorld Melting Temp Task \\
    \midrule
    \makecell[l]{
        Subgoal: The stove is turned on. on the stove is: \\
           \;\;\;\;a substance called liquid [substance].\\
        1. Focus on the thermometer\\
        2. Focus on the substance you want to heat\\
        3. Move the focused substance to the stove\\
        4. Activate the stove
    } \\
    \bottomrule
    \toprule
    \bf NetHack Task \\
    \midrule
    \makecell[l]{
    Subgoal: You succeed in unlocking the door.\\
    1. Stand adjacent to the closed door that \\
       \;\;\;\;needs to be unlocked\\
    2. Use the action 'a' to apply the relevant \\
       \;\;\;\;key or tool that can unlock the door\\
    3. Confirm the unlock action by responding \\
       \;\;\;\;affirmatively when prompted, typically \\
       \;\;\;\;by using the action 'y'
    } \\
    \bottomrule
    \end{tabular}
    \caption{
    Example generated skills.
    }
    \label{tab:skill}
\end{table}

\section{Experimental Results}
\label{sec:results}

In both ScienceWorld and NetHack domains, \methodshort{} provides large performance gains by informing the LLM actor about potential subgoals and how to accomplish them.
Figure \ref{fig:res} reports \methodshort's state-of-the-art performance for both task adaptation and transfer in ScienceWorld as well as \methodshort's superior performance and continual learning ability when compared with ReAct and Reflexion baselines in NetHack.
Error bars for NetHack show standard error across five training runs.
We do not provide standard error for ScienceWorld because the large number of tasks and expensive GPT-4 actor limited our computational budget.
In the following sections, we analyze what makes \methodshort{} successful.

\subsection{Skill Analysis}

Figure \ref{fig:stats} reports several skill statistics throughout \methodshort{} training.
\methodshort{} continuously increases the size of its skill set during training, and the rate of constructing new skills decreases as the skill set increases in experience coverage.
In our experiments, we retrieve up to three skills to include in-context at each step, but much fewer than this are self-reported as ``executed'' during the trajectory.
Despite this, our experience suggests that even unreported skills improve LLM actor performance.
Also, the number of skills being executed per trajectory steadily increases, suggesting that learned skills become more useful throughout training.
Finally, the subtrajectories used to create skills have an average length of 2.64 and 2.95 for ScienceWorld and NetHack respectively.
Remember that we limit subtrajectory length to [2, 5] in our experiments.
The length of generated instructions is slightly longer at 3.18 and 3.27 respectively.

Qualitatively, we found that \methodshort{} was especially helpful with less intuitive aspects of the action space.
For example, the ScienceWorld Melting Temperature task requires using the \textit{focus} action on the thermometer and substance before attempting to melt it, and NetHack requires using the \textit{apply} action on a key to unlock a nearby door.
Humans can quickly adapt to unintuitive domain specific requirements after a few tries, but we found that the unaltered LLM actors often got stuck in situations where actions were unintuitive outputting text such as ``I apologize for the confusion. I am trying to...'' before attempting the invalid action again.
However, skills generated by \methodshort{}, such as those in Table \ref{tab:skill}, allow the actor to reliably pass bottlenecks and continue to explore the next steps of the task.

\begin{figure*}
    \centering
    \includegraphics[width=.85\linewidth]{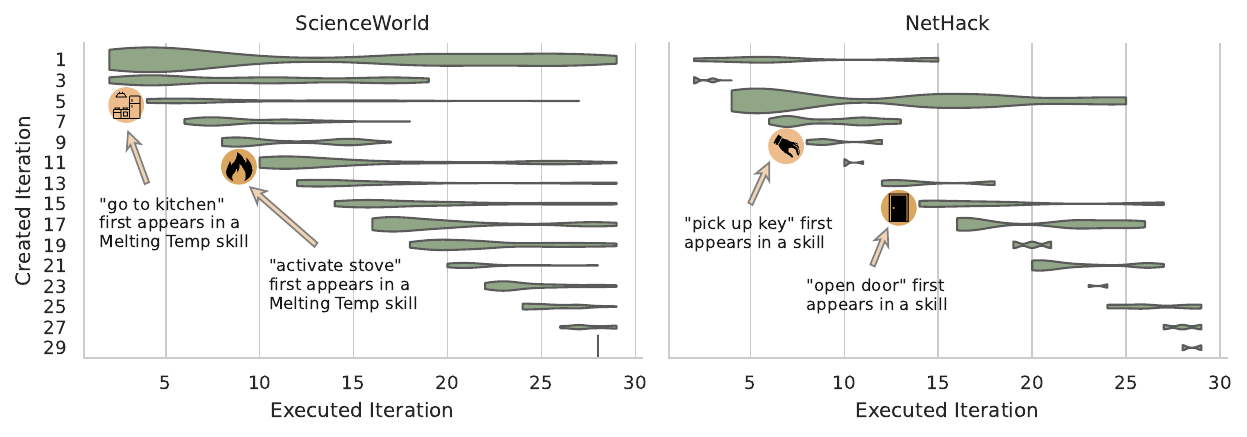}
    \caption{
    Visualization of the lifecycle of \methodshort{} skills.
    For every skill created in an iteration of \methodshort{}, the corresponding row shows when that iteration's skills were executed throughout training.
    Most skills are pruned soon after creation and replaced with improved skills.
    The icons in the figure indicate when skills with the corresponding actions were first created.
    }
    \label{fig:usage}
\end{figure*}

We hypothesize that \methodshort{} continually optimizes the LLM actor's policy by iteratively creating better skills and gathering better data.
We look for evidence of this by visualizing the skill lifecycle in Figure \ref{fig:usage}.
Each row of the figure represents the skills that were created in the corresponding iteration of \methodshort{} and when those skills were executed during training.
The densities in the figure are normalized by the total number of skills, so the the density width indicates both how many skills were created in that iteration and in which iterations they were executed.

Figure \ref{fig:usage} shows that skill set refinement prunes most skills shortly after they are created.
Also, while skills from the initial iterations are used for longer, the LLM actor tends to use skills that were created more recently, suggesting that more useful skills are discovered later in training.
This is further supported by visualizing when certain actions first appear in skills, as shown by the icons in Figure \ref{fig:usage}.
Note that the icons indicate when a skill first appears, but improved versions of that same skill may be created in later iterations.
In general, actions that are required later in a trajectory are included in skills later in training. 

\begin{figure}
    \centering
    \includegraphics[trim={0 .5cm 0 0},width=\linewidth]{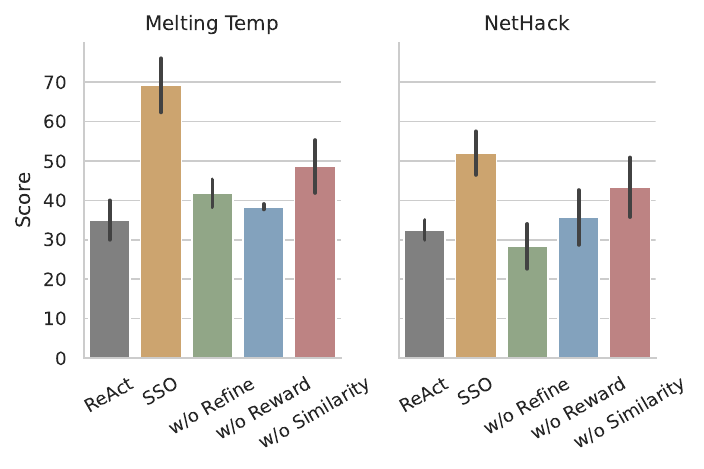}
    \caption{
    \methodshort{} ablations with and without refinement, reward weight during scoring, and similarity-based extraction.
    }
    \label{fig:ablations}
\end{figure}

\subsection{Ablations}

We compare \methodshort{} to several ablations on ScienceWorld's Melting Temperature task and the NetHack task.
First, we ablate the skill refinement method described in Section \ref{sec:refine} and do not ever prune the skill set.
Next, we ablate the use of task reward in scoring skills for sampling by reducing the corresponding score weight to zero.
Finally, we ablate similarity-based extraction and instead use the three steps before every environment reward to generate skills.

Figure \ref{fig:ablations} shows that each of these ablations decreases the performance of \methodshort{} but still regularly outperform the baseline.
Skill set refinement appears to be slightly more impactful than environment rewards and extracting pairs of similar subtrajectories.
Using similarity as an extraction method appears less impactful than simply using environment rewards.
However, including similarity-based extraction still contributes to \methodshort{} by automatically determining the start and end states for high-reward subtrajectories.

\section{Discussion \& Conclusion}

\methodlong{} (\methodshort{}) is a new in-context policy optimization method that allows LLM actors to quickly adapt to and transfer between tasks.
\methodshort{} achieves state-of-the-art results in the ScienceWorld domain and outperforms all experimented baselines on our custom NetHack task.
Every iteration, \methodshort{} constructs commonly executed high-reward skills to add to the skill set and filters out poorly performing skills.
Unlike previous work, \methodshort{} utilizes ongoing environment feedback to evaluate and prune skills, leverages intermediate rewards to identify subgoals, facilitates skill retrieval, and learns abstract transferable skills.

One limitation of \methodshort{} is the current similarity metric used for extracting potential skills.
State and action embedding similarity may be less effective in environments with distracting state information or low-level actions.
Also, although \methodshort{} is capable of operating in environments without intermediate rewards, for best performance task designers must include intermediate subgoal rewards in the environment feedback.
Fortunately, this is a common paradigm for sequential decision making.
Finally, \methodshort{} on its own does not include a method for leveraging negative environment feedback outside skill set refinement, but a method such as Reflexion \cite{shinn2023reflexion} can easily be used in addition to \methodshort{} to provide this feature.
We provide these limitations as potential inspiration for future work.

We believe that \methodshort{} takes a significant step towards reliable in-context policy optimization methods for LLM actors.
\methodshort{} manages this while being a general purpose solution, not limited by specific output formats \cite{wang2023voyager}, and capable of operating in scenarios with \cite{zhao2023expel} or without \cite{majumder2023clin} environment reinforcement signals.
We hope \methodshort{} inspires continued research in improving the effectiveness and learning efficiency of continual learning for LLM actors.

\section*{Impact Statement}

This work aims to improve the ability of deployed LLMs to continually adapt and transfer knowledge between tasks without retraining.
This research direction has the potential to improve many state-of-the-art LLM applications, but does not alter any of their fundamental limitations.

\section*{Acknowledgements}

We would like to thank the Aristo team at the Allen Institute for Artificial Intelligence for their discussions and feedback.
This work was funded in part by the DARPA ANSR program under award FA8750-23-2-0004.

\bibliography{custom}
\bibliographystyle{icml2024}

\newpage
\appendix
\onecolumn

\section{Models and Hyperparameters}

All of our ScienceWorld experiments were completed with OpenAI's \verb_gpt-4-0613_, and our NetHack experiments were completed with \verb_gpt-4-1106-preview_.
We used \verb_text-embedding-ada-002_ as the embedding model for all similarity metrics.
Table \ref{tab:params} shows the hyperparameters we used in all of our experiments.
We found that these hyperparameters we robust and worked well in all of our experiments on both domains.
However, we expect that the parameters regulating skill length and number of retrieved skills may need to be adjusted in domains with skills that operate at a different granularity.

\begin{table}[h]
    \center
    \begin{tabular}{lc}
        Parameter & Value \\
        \midrule
        Max skill length & 5 \\
        Min skill length & 2 \\
        Adaptation training episodes & 5 \\
        Transfer training episodes & 30 \\
        Sampling temp (train) & 0.7 \\
        Sampling temp (test) & 0.0 \\
        Max retrieved skills & 3 \\
        Skill refinement threshold & 0 \\
        Skill length score weight & 0.01 \\
        Reward score weight & 0.1 \\
        State similarity score weight & 1.0 \\
        Action similarity score weight & 1.0
    \end{tabular}
    \caption{\methodlong{} hyperparameters.}
    \label{tab:params}
\end{table}

\section{Prompts and Examples}

Below we detail the prompts used for our LLM actor and generating skills. All prompts are the same across domains besides task, state, and admissible action descriptions. Text marked with \verb_<>_ indicate where variable text is inserted. We also include example skills in Tables \ref{tab:sw_skills} and \ref{tab:nh_skills} and corresponding subtrajectories in Tables \ref{tab:kitchen_traj} and \ref{tab:pickup_traj}.

\subsection{LLM Actor Prompt}

\begin{verbatim}
You are playing a text-based game in which you must interact with your 
surroundings to complete a task. You will occasionally be given posisible 
subgoals. You may choose to target one of these subgoals or ignore them.

<task description>

Given the state, reflect on what has happened so far, explain your plan to 
accomplish the task, output which of the given subgoals you are targeting next 
(match one of the subgoals in the prompt word for word or output "none"), and 
then output the next action to execute (use one of the action templates below).

For example:
The last action had the effect of... To accomplish the task, I will need to...
Current subgoal: [subgoal]
Next action: [action]

<admissible actions>

The following instructions contain potentially useful information about 
reaching subgoals:

Instructions for reaching the subgoal <skill subgoal>:
    1. <skill instruction>
    2. <skill instruction>

Instructions for reaching the subgoal <skill subgoal>:
    1. <skill instruction>
    2. <skill instruction>

<state description>
\end{verbatim}

\subsection{Skill Generation Prompt}

\begin{verbatim}
You are an expert planning system. You are creating reusable skills to execute 
when completing various tasks. You create skills by looking at successful 
examples of task completions. A skill is composed of a list of instructions and 
a target state. After creating a skill, it will be used to execute actions in 
an environment. The environment will return a set of observations that 
summarize the new environment state. These observations will be used in 
conjunction with the skill's target state to determine whether the last skill 
was successful.

Consider the example trajectories of states and actions below. You'll be asked 
to analyze the similarities between each. Pay attention to the wording of the 
state observations and actions. Then you'll be asked to generate the common 
instructions, and target state for them.

Example 1:
<subtrajectory>

Example 2:
<subtrajectory>

Generate a summary of what is happening in the examples above and the 
similarities between them. Provide a name for the skill that is being executed 
in the examples above. Do not generate skill instructions or target yet.

<LLM output, unused>

Generate a numbered list of instructions for completing the skill. The 
instructions should be similar to the actions in the examples. Instructions 
should use the action templates provided below. Create generic instructions 
that would be valid for every example but specific enough to be useful in the 
examples. Do not mention the examples in the instructions. Use the output 
format:
Skill [skill name] instructions:
1. instruction 1
2. instruction 2
...

Action templates: <admissible actions>

<LLM output, skill instructions>

Generate a single target observation that would indicate the success of the 
skill. The target should be similar to one of the observations in the final 
states. Create a generic target that would be valid for every example. Do not 
mention the examples in the target. Use the output format:
Skill [skill name] target: [target observation]

<LLM output, skill subgoal>
\end{verbatim}

\newpage
\section{\methodshort{} Code}
\label{app:code}

We include the following python code as a high-level overview of \methodshort{}. However, the complete codebase is available at \href{https://github.com/allenai/sso}{https://github.com/allenai/sso}.

\begin{verbatim}
def train(env, agent, skillset, iterations=30):
    for _ in range(iterations):

        state = env.reset()
        trajectory = []
        done = False
        while not done:

            skills = skillset.retrieve(state)
            executed_skill, action = agent.act(state, skills)
            next_state, reward, done = env.step(action)

            trajectory.append((state, action, executed_skill, reward))
            state = next_state

        skillset = construct(skillset, trajectory)
        skillset = refine(skillset, trajectory)

def construct(skillset, trajectory, N=10, min=2, max=5, weights=[1, .1, .01]):
    # Extract sets of similar subtrajectories
    skillset.trajectories.append(trajectory)
    subtraj_sets = extract(skillset.trajectories[-N:], min, max)

    # Score sets of subtrajectories according to similarity, reward, and length
    scored_subtraj_sets = score(scored_subtraj_sets, weights)

    # Sample sets of subtrajectories according to score without overlapping
    sampled_subtraj_sets = sample(scored_subtraj_sets)

    # Generate skills from sets of subtrajectories
    new_skills = generate(sampled_subtraj_sets)

    skillset.skills.update(new_skills)
    return skillset

def refine(skillset, trajectory, gamma=.9, epsilon=0):
    for t in range(len(trajectory)):
        state, action, executed_skill, reward = trajectory[t]

        if executed_skill:
            discounted_reward = sum(
                trajectory[t+i][3] * gamma ** i
                for i in range(len(trajectory)-t)
            )
            skillset.observed_values[executed_skill] += discounted_reward

            if skillset.observed_values[executed_skill] <= epsilon:
                skillset.skills.remove(executed_skill)

    return skillset
\end{verbatim}

\begin{table}[p]
    \centering
    \small
    \begin{tabular}{p{0.08\linewidth} p{0.08\linewidth} p{0.08\linewidth} p{0.3\linewidth} l}
        \bf Created Iter & \bf Executed Count & \bf Observed Value & \bf Subgoal & \bf Instructions \\\midrule
        3 & 19 & 2.25 & you focus on the thermometer & \makecell[l]{1. wait\\2. focus on thermometer} \\\midrule
        5 & 22 & 5.06 & you move to the kitchen & \makecell[l]{1. go hallway\\2. go kitchen} \\\midrule
        11 & 6 & 1.28 & the stove is turned on. on the stove is: a substance called liquid [substance] & \makecell[l]{1. focus on the thermometer\\2. focus on the substance you want to heat\\3. move the focused substance to the stove\\4. activate the stove} \\\midrule
        11 & 9 & 0.37 & ``it's not clear how to read that." & \makecell[l]{1. if the thermometer is not already on the\\stove, move thermometer to stove\\2. focus on the thermometer\\3. attempt to read the thermometer\\4. wait} \\\midrule
        13 & 1 & 0.65 & you focus on the thermometer & \makecell[l]{1. if the stove is not activated, activate stove\\2. move the object (if any) to the stove\\3. focus on thermometer} \\\midrule
        24 & 1 & 0.66 & successfully read the thermometer & \makecell[l]{1. focus on thermometer\\2. read thermometer} \\\midrule
    \end{tabular}
    \caption{Top six learned skills for the Melting Temperature ScienceWorld task according to observed value at the end of training.}
    \label{tab:sw_skills}
\end{table}

\begin{table}[p]
    \centering
    \small
    \begin{tabular}{p{0.08\linewidth} p{0.08\linewidth} p{0.08\linewidth} p{0.22\linewidth} l}
    \bf Created Iter & \bf Executed Count & \bf Observed Value & \bf Subgoal & \bf Instructions \\\midrule
    9 & 2 & 0.20 & you succeed in unlocking the door & 
        \makecell[l]{
        1. a: apply/use item (choose key from the inventory)\\
        2. k: move north (to indicate the direction of the door \\to unlock)\\
        3. y: confirm action (to unlock the door when prompted)
        } \\\midrule
    10 & 2 & 0.15 & you have a key named the master key of thievery & 
        \makecell[l]{
        1. u: move northeast towards the key\\
        2. ,: pick up the key at the current location
        } \\\midrule
    16 & 1 & 0.31 & you see an open door very near northwest & 
        \makecell[l]{
        1. press 'o' to initiate the open door action\\
        2. press 'y' to specify the northwest direction for the \\open door action
        } \\\midrule
    20 & 2 & 0.26 & the door opens, or you unlock the door & 
        \makecell[l]{
        1. o - open door\\
        2. y - move or target northwest direction when prompted\\ for direction after attempting to open a door\\
        3. a - apply/use item from inventory\\
        4. g - choose the key for application when prompted\\ for the item to use (assuming 'g' correlates to the\\ key in the inventory list)\\
        5. y - move or target northwest direction when prompted\\ for direction after selecting the key to apply
        } \\\midrule
    23 & 2 & 0.18 & the door is open & 
        \makecell[l]{
        1. correctly orient towards the door if not already facing it\\
        2. use 'o' to attempt to open the door\\
        3. if the door does not open and is locked, use 'a' to apply\\ the key or lock pick to the door
        } \\\midrule
    25 & 3 & 0.29 & you have a key named the master key of thievery & 
        \makecell[l]{
        1. move to the location of the key if not already adjacent\\ (using k, l, j, h, y, u, b, or n as appropriate)\\
        2. identify the key at the current location ('y' to move\\ northwest if the key is northwest of the agent)\\
        3. pick up the key at the current location (',')
        } \\\midrule
    \end{tabular}
    \caption{Top six learned skills for the NetHack task according to observed value at the end of training.}
    \label{tab:nh_skills}
\end{table}

\begin{table}[p]
    \centering
    \begin{tabular}{cc}
        \bf Subtraj 1 & \bf Subtraj 2 \\
        \makecell*[{{p{.45\linewidth}}}]{
        Initial State: \\
        This room is called the art studio. In it, you see:; the agent; a substance called air; a large cupboard. The large cupboard door is closed.; a table. On the table is: a jug (containing nothing).; a wood cup (containing yellow paint); a wood cup (containing red paint); a wood cup (containing blue paint); You also see: A door to the hallway (that is open); In your inventory, you see: an orange \\\\
        Trajectory: \\
        Action: go hallway \\
        Observation: You move to the hallway\\
        Action: focus go kitchen \\
        Observation: You move to the kitchen\\\\
        Final State: \\
        You move to the kitchen.; This room is called the kitchen. In it, you see:; the agent; a substance called air; a ceramic cup (containing a substance called lead); a chair. On the chair is: nothing.; a counter. On the counter is: a bowl (containing a red apple, a banana, an orange, a potato), a drawer.; a cupboard. The cupboard door is closed.; a freezer. The freezer door is closed.; a fridge. The fridge door is closed.; a glass jar (containing a substance called sodium chloride); a lighter; a orange box (containing nothing); a oven, which is turned off. The oven door is closed.; a painting; a sink, which is turned off. In the sink is: nothing.; a substance called soap; a stopwatch, which is deactivated.; a stove, which is turned off. On the stove is: nothing.; a table. On the table is: a glass cup (containing nothing).; a thermometer, currently reading a temperature of 10 degrees celsius; a yellow box (containing nothing); You also see: A door to the bathroom (that is open), A door to the hallway (that is open), A door to the outside (that is open); In your inventory, you see: an orange
        } &
        \makecell*[{{p{.45\linewidth}}}]{
        Initial State: \\
        This room is called the greenhouse. In it, you see:; the agent; a substance called air; a bee hive. The bee hive door is closed.; a jug (containing nothing); a shovel; a sink, which is turned off. In the sink is: nothing.; You also see: A door to the hallway (that is open), A door to the outside (that is open); In your inventory, you see: an orange \\\\
        Trajectory: \\
        Action: go hallway \\
        Observation: You move to the hallway\\
        Action: focus go kitchen \\
        Observation: You move to the kitchen\\\\
        Final State: \\
        You move to the kitchen.; This room is called the kitchen. In it, you see:; the agent; a substance called air; a chair. On the chair is: nothing.; a counter. On the counter is: a bowl (containing a red apple, a banana, an orange, a potato), a drawer.; a cupboard. The cupboard door is closed.; a freezer. The freezer door is closed.; a fridge. The fridge door is closed.; a glass jar (containing a substance called sodium chloride); a lighter; a orange box (containing nothing); a oven, which is turned off. The oven door is closed.; a painting; a plastic cup (containing a substance called gallium); a sink, which is turned off. In the sink is: nothing.; a substance called soap; a stopwatch, which is deactivated.; a stove, which is turned off. On the stove is: nothing.; a table. On the table is: a glass cup (containing nothing).; a thermometer, currently reading a temperature of 10 degrees celsius; a yellow box (containing nothing); You also see: A door to the bathroom (that is open), A door to the hallway (that is open), A door to the outside (that is open); In your inventory, you see: an orange
        }
        \\
    \end{tabular}
    \caption{Subtrajectories used to generate the ``you move to the kitchen'' skill described in Table \ref{tab:sw_skills}.}
    \label{tab:kitchen_traj}
\end{table}

\begin{table}[p]
    \centering
    \begin{tabular}{cc}
        \bf Subtraj 1 & \bf Subtraj 2 \\
        \makecell*[{{p{.45\linewidth}}}]{
        Initial State: \\
        You have a +0 short sword (weapon in hand). You have 15 +0 daggers (alternate weapon; not wielded). You have an uncursed +1 leather armor (being worn). You have an uncursed potion of sickness. You have an uncursed lock pick. You have an empty uncursed sack. You see a vertical wall far east. You see a horizontal wall near north and northeast. You see a area of lava near northeast. You see a stairs down near northeast. You see a vertical wall near west. You see a horizontal closed door near northwest. You see a dark area near northwest. You see a lava very near northeast, northeast, and east. You see a horizontal wall adjacent southeast, south, and southwest. You see a key adjacent northwest. Hello Agent, welcome to NetHack!  You are a chaotic male human Rogue. \\\\
        Trajectory: \\
        Action: y \\
        Observation: You see here a key named The Master Key of Thievery\\
        Action: , \\
        Observation: g - a key named The Master Key of Thievery\\\\
        Final State: \\
        You have a +0 short sword (weapon in hand). You have 15 +0 daggers (alternate weapon; not wielded). You have an uncursed +1 leather armor (being worn). You have an uncursed potion of sickness. You have an uncursed lock pick. You have an empty uncursed sack. You have a key named The Master Key of Thievery. You see a vertical wall far east. You see a horizontal wall near north and northeast. You see a lava near northeast, east, and southeast. You see a area of lava near northeast. You see a stairs down near east. You see a vertical wall near west. You see a dark area near northwest. You see a horizontal wall very near southeast, south, and southwest. You see a horizontal closed door very near northwest. You see a stairs up adjacent southeast. g - a key named The Master Key of Thievery.
        } &
        \makecell*[{{p{.45\linewidth}}}]{
        Initial State: \\
        You have a +0 short sword (weapon in hand). You have 8 +0 daggers (alternate weapon; not wielded). You have an uncursed +1 leather armor (being worn). You have an uncursed potion of sickness. You have an uncursed lock pick. You have an empty uncursed sack. You have an uncursed blindfold. You see a vertical wall far east. You see a horizontal wall near north and northeast. You see a area of lava near northeast. You see a stairs down near northeast. You see a vertical wall near west. You see a horizontal closed door near northwest. You see a dark area near northwest. You see a lava very near northeast, northeast, and east. You see a horizontal wall adjacent southeast, south, and southwest. You see a key adjacent northwest. Hello Agent, welcome to NetHack!  You are a chaotic male human Rogue. \\\\
        Trajectory: \\
        Action: y \\
        Observation: You see here a key named The Master Key of Thievery\\
        Action: , \\
        Observation: h - a key named The Master Key of Thievery\\\\
        Final State: \\
        You have a +0 short sword (weapon in hand). You have 8 +0 daggers (alternate weapon; not wielded). You have an uncursed +1 leather armor (being worn). You have an uncursed potion of sickness. You have an uncursed lock pick. You have an empty uncursed sack. You have an uncursed blindfold. You have a key named The Master Key of Thievery. You see a stairs down far northeast. You see a vertical wall far east. You see a horizontal wall near north and northeast. You see a lava near northeast, east, and southeast. You see a area of lava near northeast. You see a vertical wall near west. You see a dark area near northwest. You see a horizontal wall very near southeast, south, and southwest. You see a horizontal closed door very near northwest. You see a stairs up adjacent southeast. h - a key named The Master Key of Thievery.
        }
        \\
    \end{tabular}
    \caption{Subtrajectories used to generate the ``you have a key named the master key of thievery'' skill described in Table \ref{tab:nh_skills}.}
    \label{tab:pickup_traj}
\end{table}

\end{document}